\documentclass[conference]{IEEEtran}
\IEEEoverridecommandlockouts
\usepackage{cite}
\usepackage{amsmath,amssymb,amsfonts}
\usepackage{graphicx}
\usepackage{textcomp}
\usepackage{xcolor}
\usepackage{graphicx}
\usepackage{multirow}
\usepackage{xcolor}
\usepackage{caption}
\usepackage{subcaption}
\usepackage{amsmath}
\usepackage{amssymb}
\usepackage{tabularx}
\usepackage{algorithm}
\usepackage{algpseudocode}
\usepackage{url}
\usepackage[justification=centering]{caption}
\graphicspath{ {./Images/} }

\makeatletter
\newcommand{\linebreakand}{%
  \end{@IEEEauthorhalign}
  \hfill\mbox{}\par
  \mbox{}\hfill\begin{@IEEEauthorhalign}
}
\makeatother
\def\BibTeX{{\rm B\kern-.05em{\sc i\kern-.025em b}\kern-.08emT\kern-.1667em\lower.7ex\hbox{E}\kern-.125emX}}

\author{
\IEEEauthorblockN{Sadia Kamal}
\IEEEauthorblockA{\textit{Department of Computer Science} \\
\textit{Oklahoma State University}\\
sadia.kamal@okstate.edu}
\and
\IEEEauthorblockN{Jimmy Hartford}
\IEEEauthorblockA{\textit{Department of Mathematics} \\
\textit{Cushing Public Schools}\\
jimmy.hartford@cushingtigers.com,}
\and
\IEEEauthorblockN{Jeremy Willis}
\IEEEauthorblockA{\textit{Department of Science \& Technology} \\
\textit{Sand Springs Public Schools}\\
jeremy.willis@sandites.org}
\linebreakand
\centering
\IEEEauthorblockN{\centering Arunkumar Bagavathi}
\IEEEauthorblockA{\textit{Department of Computer Science} \\
\textit{Oklahoma State University}\\
abagava@okstate.edu}
}

\begin{document}


\title{Learning Unbiased News Article Representations: \\ A Knowledge-Infused Approach*
\thanks{*This paper is partly supported by NSF: Research Experience for Teachers (RET); Award Number:2055557}
}
\maketitle


\begin{abstract}


Quantification of the political leaning of online news articles can aid in understanding the dynamics of political ideology in social groups and measures to  mitigating them. However, predicting the accurate political leaning of a news article with machine learning models is a challenging task. This is due to (i) the political ideology of a news article is defined by several factors, and (ii) the innate nature of existing learning models to be biased with the political bias of the news publisher during the model training. There is only a limited number of methods to study the political leaning of news articles which also do not consider the algorithmic political bias which lowers the generalization of machine learning models to predict the political leaning of news articles published by any new news publishers. In this work, we propose a knowledge-infused deep learning model that utilizes relatively reliable external data resources to learn unbiased representations of news articles using their global and local contexts. We evaluate the proposed model by setting the data in such a way that news domains or news publishers in the test set are completely unseen during the training phase. With this setup we show that the proposed model mitigates algorithmic political bias and outperforms baseline methods to predict the political leaning of news articles with up to $73\%$ accuracy. 

\end{abstract}


\section{Introduction}
\begin{figure}[h]
\centering
\includegraphics[scale=0.3]{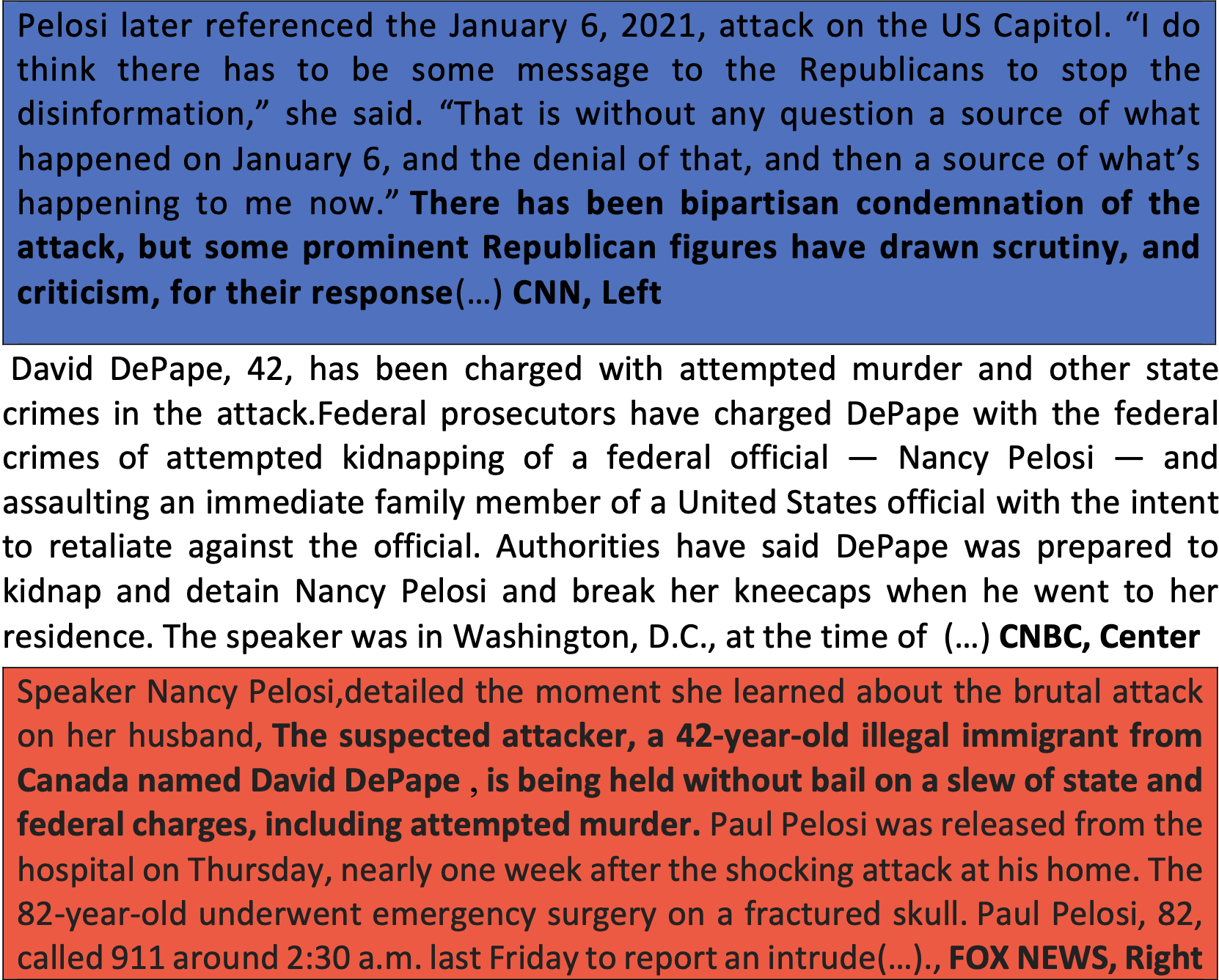}
\caption{Biased projections of news contents on the same story of 'Nancy Pelosi Discusses Attack on Husband' covered by news media houses with political leaning of \emph{left} (blue), \emph{center} (white), and \emph{right} (red)}
\label{fig:news_snippets}
\end{figure}

News media houses have endured through time to disseminate political news to the people while also influencing their political perceptions. 
They play a vital role to impersonate public opinion on several issues like diseases~\cite{alafnan2020covid}, elections~\cite{beckers2020voice}, and natural calamities~\cite{ghassabi2015role}.  Recent events from COVID-19 prove that news media bias is capable of polarizing individual ideologies~\cite{alafnan2020covid,chipidza2021effect}.
Moreover, the rise of the web and social media has also increased the capacity of news media to propagate information at a rapid pace. 
Over the last decade, several small-scale and focused online news forums have harnessed online platforms to spread news content across the general public.  
News articles reflect the news forums' opinion on politicians, laws, and policies which in turn defines the \emph{political bias} of a news media house. A news forum is politically biased in the scale of \emph{far-left} to \emph{far-right} based on the political ideology (republican/democrat in the USA) that they approve and criticize in their news articles. 
An example of news domains from different political leaning covering the story of 'attack on Mr. Pelosi' news from multiple angles is given in Figure~\ref{fig:news_snippets}. 
\emph{Right-leaning} news domains attribute the problem of illegal immigrants, \emph{left-leaning} domains emphasize Mrs.Pelosi's criticisms on republicans, and \emph{center-leaning} news domains cover details of the attack at a high-level.

Politically biased news contents are some of the well known sources for the widespread increase in filter bubbles~\cite{liu2021interaction}, echo-chambers~\cite{garimella2018political}, misinformation~\cite{shu2019beyond}, and propaganda in online communities~\cite{osmundsen2021partisan}.
Our primary motivation in this work is echo-chambers created by biased news recommendations in online forums~\cite{pennycook2019fighting,huszar2022algorithmic}. In other words, user profiles with conservative ideology would prefer news articles from right leaning news media like Fox News and CBN, whereas profiles with liberal ideology may prefer news from left leaning news domains like CNN and CBS News. 
Some of the current AI based recommendation models intensify the echo chamber effect by giving news article suggestions to users that align with their ideology and preferences. Thus there are works to refine such recommender models to expose online users to news articles from news domains of diverse political ideologies~\cite{gillani2018me,levy2021social,jeon2021chamberbreaker}.
Thus characterizing political bias in news articles is crucial to quantify its effects and take measures to mitigate them in online forums like Twitter,  Reddit, and Youtube. Accurately mapping the bias of news articles will also provide social media platforms and news recommendation platforms like Google News to show personalized news from multiple political ideologies~\cite{baly2020detect}.


Algorithmic bias is one of the primary concerns in existing real-world AI applications. The machine learning models by default undergo algorithmic bias as the training data given to such models are collected on human assumptions~\cite{baeza2016data,mehrabi2021survey}. Although these biased models give good performance on new data that have similar characteristics to the training set, the models discriminate features that are completely unseen during the training phase. Such model biases are prevalent in many real-world scenarios like gender, race, and religion~\cite{solaiman2021process}. Such unfairness in machine learning models reduces the human reliability of model outcomes and their decision-making capabilities in various domains like medicine, agriculture, and autonomous systems. In this paper, we analyze machine learning classifiers undergoing \emph{algorithmic political bias} of news domains that are available during model training and make discriminatory predictions. We demonstrate in upcoming sections that existing language models are politically biased on news domains and fail to map the political leaning of news articles published by news domains that are unseen during the training phase. It is important to mitigate such bias in the learning models to (i) learn unbiased feature representations of news articles, and (ii) predict political bias of news articles from news media sources whose bias is not evidenced during model training.

The existing work~\cite{baly2020detect} showcases the effect of political bias present in news domains to predict the political leaning of news articles. Mitigating such political bias in news articles is important to understand the political bias of a news article from any new news domain. We introduce such a bias mitigation method in a machine learning setup to learn news article representations by infusing external knowledge. Our method is an improvement over all the prior work on political leaning prediction of news articles, as summarized in Table~\ref{tab:comparison}. The proposed method of encoding features of topics learned from external data resources can enable the machine learning models to mitigate political bias in classification tasks. Overall, we have the following \emph{three-fold} contributions in this paper:


\begin{enumerate}
	\item \textbf{Proposed work:} We propose a novel framework that mitigates political bias in machine learning classifier models by infusing external knowledge sources for political leaning prediction on news articles. Most importantly, the proposed model uses representations of topics in news articles and knowledge of news domains from external data sources in the prediction task
	\item \textbf{Experiments:} We give in-depth experiments on the performance of the proposed framework using multiple training and test sets, multiple language models and external knowledge sources. We show that the proposed model achieves up to \emph{73\%} prediction accuracy
    \item \textbf{Mitigating political bias:} We demonstrate the efficacy of the proposed framework to mitigate political bias of news domains even when they are not available during the model training. Specifically, we demonstrate robustness by evaluating the proposed model with multiple splits of training and test data based on news domains in our experiments
\end{enumerate}

\begin{table}[h!]
\centering
\caption{Qualitative comparison of aspects in the proposed work with existing methods} \label{table:comparison} 

\begin{center}
\begin{tabular}{p{2.3cm}| p{1.1cm}|p{1.1cm}|p{1.1cm}|p{1.1cm}} 
 \hline
 \hline
 Method & Knowledge Encoding & Multiple Data Sources & Topic Encoding& Bias Mitigation \\ 
 \hline
 Baly et al.~\cite{baly2020detect}& \checkmark & - & -& \checkmark  \\ 

 KGAP ~\cite{https://doi.org/10.48550/arxiv.2108.03861} & \checkmark & - & \checkmark & - \\

 KCD ~\cite{Zhang2022KCDKW} & \checkmark & - & \checkmark & - \\

 Devatine et al.~\cite{devatine-etal-2022-predicting} &-&-&-&\checkmark\\

 KHAN~\cite{ko2023khan} & \checkmark & \checkmark & \checkmark& - \\
 \hline
 \textbf{Ours} & \checkmark  & \checkmark  & \checkmark &\checkmark  \\ 
 \hline
\end{tabular}
\end{center}
\label{tab:comparison}
\end{table}

 \section{Related Work}

\subsection{Political Leaning Detection}
Identifying political leaning has several applications in online social media like echo chamber detection~\cite{barbera2015tweeting}, quantifying controversy ~\cite{garimella2018quantifying}, and partisan segregation identification~\cite{conover2011political}.
Several research ideas are emerging to utilize linguistic and temporal features that exist in online content to characterize political bias and leaning. Few recent works ~\cite{stefanov-etal-2020-predicting,baly-etal-2020-written} study the identification of political ideology of media houses from social media networks like Twitter. 
Analyzing text data with deep learning methods have been widely studied to capture political viewpoints. Methods like attention mechanism~\cite{li2021using} have been utilized with deep language encoder models on contextual information to capture political perspective. 
An attention-based multi-view model is proposed in~\cite{kulkarni-etal-2018-multi} to identify the ideology of news articles from the article details such as title, content and link structure. 
Opinion-aware knowledge graph~\cite{chen2017opinion} has been proven efficient specially for graph based approaches to infer and predict the ideology. MEAN~\cite{li-goldwasser-2021-mean} detects bias in terms of entities that use entity information as external knowledge and leverage that information in multiple ways to capture political perspective. KCD ~\cite{Zhang2022KCDKW} also detects political perspective with knowledge reasoning for graph level representation learning methods. Political bias detection has been studied at multiple levels like word-level, sentence-level~\cite{bhatia-p-2018-topic}, news article-level ~\cite{gangula-etal-2019-detecting,10.1145/3184558.3186987} and news domain level~\cite{baly-etal-2019-multi}. Few recent works~\cite{kulkarni-etal-2018-multi,baly2020detect,article} try to infer the political bias associated with the news article by using multiple methods such as deep neural network architectures. Although the work proposed in ~\cite{baly2020detect} is closely similar to our work, there are major differences. 
In this paper, we develop a novel knowledge-infused deep learning model to extract news article representations to predict their political leaning. We use two external sources to add both global and fine-grained contextual information to the news articles.


\subsection{Model Fairness}
Traditional machine learning and deep learning models are mostly designed to optimize based on any performance metric such as accuracy, which can lead to learning bias or unfairness~\cite{mehrabi2021survey} to the model which can affect the prediction task. Previously, there has been work like elimination and measuring of gender bias~\cite{Bordia2019IdentifyingAR}, neuro-imaging dataset bias~\cite{DINSDALE2021117689} from machine learning models and deep neural network frameworks. 
The bias in the texts associated with the sources~\cite{Baly2018PredictingFO} and users can be directly attributed to those sources and users. There are few works that try to detect the media bias, particularly the work given in ~\cite{10.1145/3308560.3316460} predicts the media bias by link-based approach. 
Having articles from the same news domain can make the model biased ~\cite{chen-etal-2020-detecting} because the classifiers may start making predictions based on the political leaning of news domain itself rather than the content of the article. We attempt to mitigate such model's algorithmic political bias in our work by \emph{media-based split} datasets and a proposed knowledge infusion method.

\section{Datasets}
In this work, we utilize datasets from multiple resources for knowledge infusion on news article representations. In this section, we give details of all the datasets.




\begin{table}[htbp]
    \centering
    \caption{Demonstration of algorithmic political bias in classifier models and the importance of external data sources - Wikipedia}
    \begin{tabular}{c|c c c}
        \hline
        \hline
         \textbf{Dataset} & \textbf{Acc.} & \textbf{Macro F1} & \textbf{MAE} \\
         \hline
         NewsA (Random split) & 0.749 & 0.7456 & 0.3774\\ 
         NewsA (Media split) & 0.516 & 0.4804 & 0.5907\\
         \hline
         WikiA & 0.562 & 0.4040 & 0.5625 \\
         NewsA (Media split) + WikiA & 0.685 & 0.6290 & 0.6011 \\
         \hline
    \end{tabular}
    \label{tab:political_bias}
\end{table}

\subsection{Labeled News Articles}
We utilize labeled news articles from the existing research~\cite{baly2020detect} in our experiments. This dataset comprises \emph{37,554} news articles from \emph{389} news domains and each news article is associated with one of the three political leaning labels: \emph{left}, \emph{center}, and \emph{right}. Prediction models trained using the traditional approach of randomly splitting the data into training and test datasets will be unfair. Meaning that such models cannot generalize to predict the political leaning of news articles from an unknown news domain, as given in \emph{row 2} (Madia split) of Table~\ref{tab:political_bias}. 



\begin{figure}
     \centering
     \begin{subfigure}[b]{0.45\textwidth}
         \centering
         \includegraphics[scale=0.52]{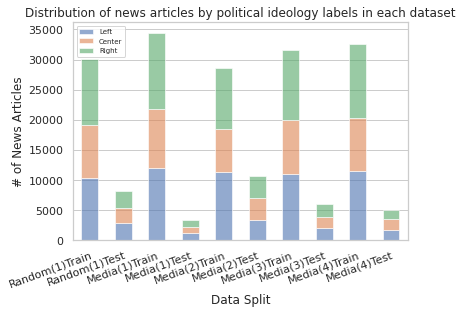}
         \caption{Distribution of political leaning labels training and test sets in one random split and four media-based data splits}
         \label{fig:data_splits}
     \end{subfigure}
     \hfill
     \begin{subfigure}[b]{0.5\textwidth}
         \centering
         \includegraphics[scale=0.55]{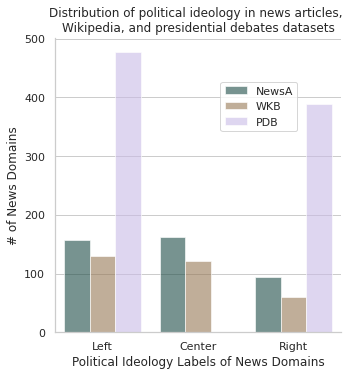}
         \caption{Distribution on the number of news domains in the news articles dataset, the number of Wikipedia articles and the number of speech in presidential debates in \emph{left}, \emph{center}, and \emph{right} ideologies}
         \label{fig:three sin x}
     \end{subfigure}
        \caption{Data distribution of news articles, news domains, Wikipedia articles, and Presidential debates characterized by their political ideologies}
        \label{fig:domain_splits}
\end{figure}

To analyze and mitigate the model bias of learning models, we split the news articles by news domains into training and test sets. Such data splits are represented as \emph{Media split} throughout this paper and this data split will test the performance of classifier models to predict the political leaning of news articles from completely unknown news domains during the model testing phase. Unlike the existing work~\cite{baly2020detect}, which hand pick the news domains for the test set we randomly choose $7\%$ of the news domains from our data and their corresponding news articles for our test set. Table~\ref{tab:political_bias} demonstrates the performance of a classifier model built on top of a BERT language model to predict the political bias of news articles. Our experiments aligns with the existing work~\cite{baly2020detect}, although ours perform better than their results, to show that the learning models underperform when they get articles from unknown news domains and it requires mitigation measures. The distribution of 4 \emph{Media split} data and one \emph{random split} data used in our experiments is given in Figure~\ref{fig:data_splits}. It is notable from the Figure~\ref{fig:three sin x} that the number of news domains in our news articles is relatively minimum. We claim that this imbalanced distribution does not bias the proposed model as the number of articles in the \emph{right} political ideology is approximately equal to the number of news articles in other political ideologies from Figure~\ref{fig:data_splits}.



\begin{figure*}[htbp]
\centering
\includegraphics[scale=0.42]{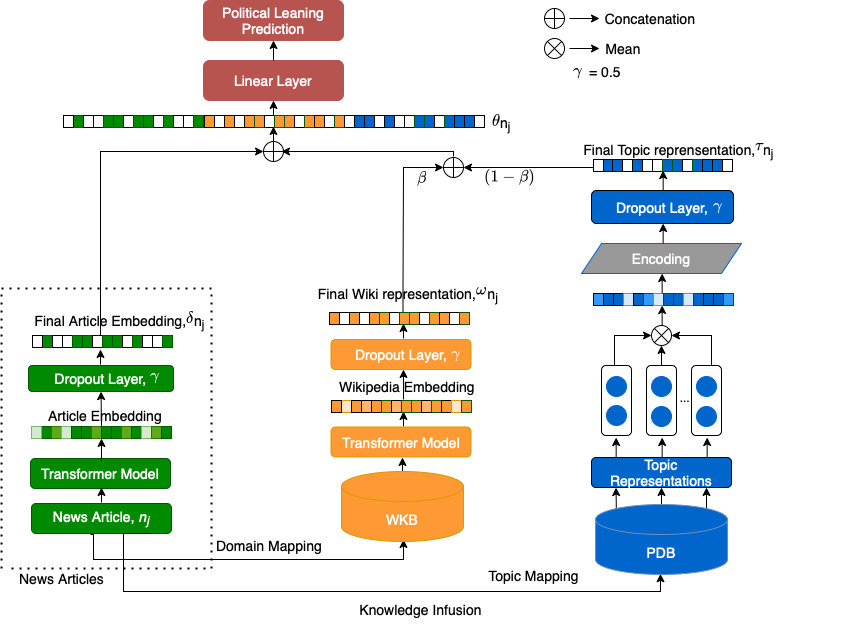}
\caption{Architecture of our proposed knowledge-infused deep learning model. The proposed model infuse weighted Wikipedia representations ($\omega$) and topics representations ($\tau$) to the corresponding base representations ($\delta$) of a news article $n_j$}
\label{fig:model_a}
\end{figure*}

\subsection{External Data Sources}
It is well studied research that adding contextual information from external data sources improve the performance of supervised machine learning models~\cite{baly2020detect,baly-etal-2020-written,farimani2021leveraging,ko2023khan}. Our proposed method use two external datasets to mitigate algorithmic political bias in political leaning prediction models. We have released the datasets to the research community here\footnote{https://github.com/sadiakamal/Learning-Unbiased-News-Article-Representation}. Figure~\ref{fig:three sin x} gives the distribution of the number of documents per class label.


\subsubsection{Wikipedia Articles (\textbf{WKB})}
One of the datasets that we use to mitigate model bias are Wikipedia pages of news domains. We use the Wikipedia API to collect the title and body of \emph{324} news domain that exists in our dataset. There are no wikipedia pages for the other 65 news domains. We used wikipedia articles in the proposed model to capture global context of news articles in the form of news domains. The observations in \emph{row 3} of Table~\ref{tab:political_bias} clearly demonstrates that Wikipedia article representations, learned from a pre-trained BERT model without incorporating any news articles representations, can effectively mitigate algorithmic political bias in the prediction model. This approach resulted in a $5\%$ improve in performance when evaluated with the test data of the \emph{media split}. We can also notice a significant performance spike of $68.5\%$ model accuracy when using Wikipedia representations along with news article representations on the media split. 

\subsubsection{Presidential Debates (\textbf{PDB})}
In addition to Wikipedia articles, we use presidential debates~\cite{ontheissues} to capture contextual representations of topics that are present in news articles. This dataset consists of debates of presidential candidates from both the republican and democratic parties. The main motivation to utilize this dataset is two-fold: (i) to capture the fine-grained representations of topics that appear in news articles, and (ii) to rely on relatively reliable external knowledge sources instead of utilizing topic representations derived from much unreliable and noisy data sources such as social media posts~\cite{baly2020detect,baly-etal-2020-written,ko2023khan}. As depicted in Figure~\ref{fig:three sin x}, our presidential debates dataset (PDB) comprises of an almost balanced distribution of 478 speeches of Democrats and 389 speeches of Republicans. We also highlight that the proposed approach extracts topic representations from the entire debate data rather than handling them separately by their political ideology. So, the absence of speeches from the \emph{center} ideology and the presence of a minor imbalance in the debates data cannot bias the proposed model. 



\section{Methodology}
\label{sect:Method}

Our proposed framework is a knowledge infused deep network architecture, as illustrated in Figure~\ref{fig:model_a}, that mitigates algorithmic political bias that exists in classifier models to predict the political leaning of news articles. Also, we summarize the list of notations used in this paper in Table~\ref{tab:notations}.

\begin{table}[h]
\caption{Description of notations used in this paper} \label{table:abl_3} 

\centering
\begin{tabular}{c  p{6cm}}
 \hline
 \hline
\textbf{Notation} & \textbf{Description} \\
 \hline

$\mathcal{N}$ & Set of news articles \\
l & Number of news articles \\
$\mathcal{D}$ & Set of news domains \\
m & Number of news domains \\
$T_{n_j}$ & Set of topics in a news article $n_j$ \\
$C_{n_j}$ & Political leaning of news article $n_j$ \\
\hline

$\delta$ & Feature representation of news articles \\
$\omega$ & Feature representation of wikipedia articles \\
$q$ & Number of dimensions in $\delta$  and  $\omega$ \\
$\tau$ & Feature representation of topics \\
$r$ & Number of dimensions in $\tau$ \\
\hline

$\beta$ & Weight of external knowledge representations \\
$\lambda$ & Weighted external knowledge representations \\
$\Theta$ & Knowledge infused news article representations \\
p & Number of dimensions in $\Theta$
\end{tabular}
\label{tab:notations}
\end{table}


Assume the training set consists of $l$ news articles which are represented as $\mathcal{N} = \{n_{1},n_{2},n_{3},...,n_{l}\}$, where each article $n_j \in \mathcal{N}$ is published by a news domain $d_k \in \mathcal{D}$ where $\mathcal{D}= \{d_{1},d_{2},d_{3},...,d_{m}\}$, given that $m<<<l$, and $n_j \in \mathcal{N}$ is associated to one of the political leaning classes $\mathcal{C}_{n_j} \in \{0,1,2\}$ (0: Left, 1: Center, 2: Right). Each news article $n_j \in \mathcal{N}$ is in turn consists of a series of topics represented as $T_{n_j} = \{t_1, t_2, t_3, \ldots\}$, given that our model does not set any limits on the number of topics in the news article $n_j$.
In addition to the news articles, we also have Wikipedia knowledge base (\textbf{WKB}) with $m$ wikipedia pages and debates database (\textbf{PDB}). The WKB provides contextual information about news domains that publish news articles $\mathcal{N}$, while the PDB focuses on the topics discussed in news articles $\mathcal{N}$.

Given a set of news articles $\mathcal{N}$ and their corresponding external data sources, the proposed model learns a $p-$dimensional representation $\Theta_{n_j} \in \mathbb{R}^{l \times p}$ of a news article $n_j$ where $p = 2q + r$. We learn the news article representation as $\Theta_{n_j} = \delta_{n_j} \oplus \omega_{n_j} \oplus \tau_{n_j}$, where $\delta_{n_j} \in \mathbb{R}^{l \times q}$ is the base representation of $n_j$, $\omega_{n_j} \in \mathbb{R}^{l \times q}$ is the representation of a Wikipedia article that corresponds to the news domain $d_k$ which published $n_j$, and $\tau_{n_j} \in \mathbb{R}^{l \times r}$ is the aggregated representation of topics in $n_j$. The proposed model learns all the above representations jointly in a supervised fashion as $f(\Theta) \rightarrow Pr(\mathcal{C}_{n_j}|\Theta)$ where $\mathcal{C}_{n_j} \in$ $\{0,1,2\}$ is the political leaning of news article $n_j \in \mathcal{N}$. We give details about the proposed model to learn $\Theta_{n_j}$, $\delta_{n_j}$, $\omega_{n_j}$, and $\tau_{n_j}$ below.



\subsection{Base Representation $\delta_{n_j}$}
We use the backbone transformer models to extract the base news article representations $\delta_{n_j} \in \mathbb{R}^{l \times q}$. We choose transformer models over classic text representation models because of its ability to learn contextualized representations rather than static word embeddings.
Given that there are $w$ words in a news article $n_j$, we aggregate the word representations from the transformer model using a MEAN operator. We fine-tune the parameters of the backbone model in a supervised setup $f(\Theta)$ as mentioned before.


\begin{algorithm}
\begin{algorithmic}[1]
\caption{Topic Knowledge Extraction}\label{alg:cap}
\Require Topics of news article $n_j \in N$: $T_{n_j}$, Pre-trained word embedding model: $\mathcal{W}$, knowledge weight variable: $\beta$
\State Initialize $E \gets \phi$
\For {topic $t \in T_{n_j}$}
\State $ E \gets E + \mathcal{W}[t] $
\EndFor
\State $ M \gets \frac{E}{|T_{n_j}|}$  \Comment{Mean of all word vectors}
\State $\tau_{n_j} \gets $ Encode($M$) to $r$-dimensional vector
\State $\tau_{n_j} \gets (1-\beta) \times \tau_{n_j}$ \Comment{weighted topic representation of $n_j$}
\State \Return {$\tau_{n_j}$}
\label{alg:topic_representation}
\end{algorithmic}
\end{algorithm}

\subsection{Knowledge Infusion}
We further enhance the base representations of news articles ($\delta_{n_j}$) with the global contextual representations of news articles ($\lambda_{n_j}$) collected from external data sources to debias the learning model $f(\Theta)$.
In this paper, we utilize Wikipedia data (\textbf{WKB}) and Presidential debates data (\textbf{PDB}) to learn the global representations $\lambda_{n_j}$ of a news article $n_j \in \mathcal{N}$. Each of our external data covers a different context of global representations. That is the representations $\omega_{n_j}$ from \textbf{WKB} consists of global context of the news domain $d_{n_j}$ and the representations $\tau_{n_j}$ from \textbf{PDB} captures the unmanipulated context of topics $T_{n_j}$ present in news articles. We primarily use political debates to learn topic representations, rather than general sources like social media posts or Wikipedia, as they are significant to the political polarization problem and they are relatively reliable to perceive the political context of topics.


\subsubsection{Wikipedia Representation $\omega_{n_j}$}
We first map the news domain $\mathcal{D}_{n_j}$ of a news article $n_j$ to its corresponding Wikipedia article $W_{n_j}$ from the Wikipedia data base \textbf{WKB} using the mapping function $g_{wkb}: \mathcal{D}_{n_j} \rightarrow W_{n_j}$.
We learn the Wikipedia representations $\omega_{n_j}$ of news article $n_j$ with the same backbone transformer model used to learn the base representations of $n_j$. 
The proposed model optimizes the backbone transformer model parameters based on the classification task $f(\Theta)$ to learn $\omega_{n_j}$.

\begin{figure}
     \centering

     \hfill
     \begin{subfigure}[b]{0.45\textwidth}
         \centering
         \includegraphics[width=\textwidth]{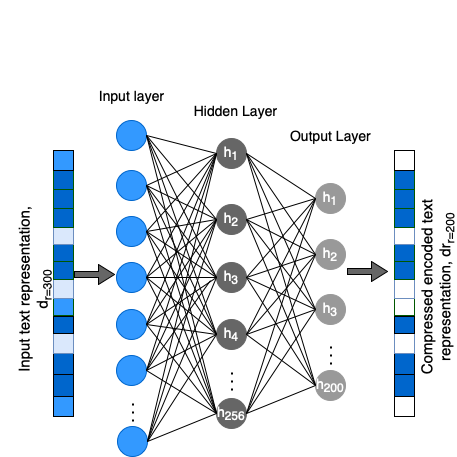}
         \caption{The Encoder model receives r=300-dimensional input text representation and maps it to a lower dimensional (r=200)representation}
         \label{fig:encoder}
     \end{subfigure}
     \hfill
          \begin{subfigure}[b]{0.45\textwidth}
         \centering
         \includegraphics[width=\textwidth]{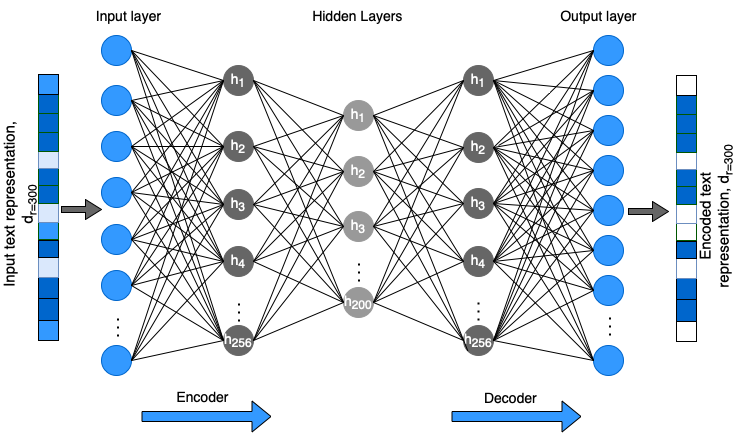}
         \caption{The Autoencoder consists of two main components: an encoder and a decoder. The encoder part works the same. The decoder part then reconstructs the input to the original 300-dimensional representation.}
         \label{fig:autoencoder}
     \end{subfigure}
        \caption{We using an Autoencoder and Encoder model to learn topic representations $\tau_{n_j}$, where each $\tau_{n_j}$ represents a topic. Autoencoder model to learn topic representations $\tau_{n_j}$ and we use ReLU() as the activation function in the hidden layers. }
        \label{fig:topic_representations}
\end{figure}

\subsubsection{Topic representation $\tau_{n_j}$}
Algorithm~\ref{alg:topic_representation} gives the complete pipeline to extract the topics representation $\tau_{n_j}$ of a news article $n_j$ with topics $T_{n_j} = \{t_1,t_2,t_3,\ldots\}$ from our presidential speeches knowledge base (\textbf{PDB}). We collect topic representation $\tau_{n_j}$ for each base representation $\delta_{n_j}$ and this is our another infused knowledge base to enhance the learning. We do not set any limits on the number of topics to appear in the news article. Capturing this information allow us to predict the political leaning of a article with political speech based topic representation learning context. First, we pre-train a skip-gram word embedding model $\mathcal{W}$ (\emph{Word2Vec}~\cite{NIPS2013_9aa42b31} in this paper) on the \textbf{PDB} data to get contextual text features of words in the context of \textbf{PDB}. We then collect representations for each topic $t \in T_{n_j}$  using the skip-gram model and take mean of representations from all topics in news article $n_j$. We further enhance topics representation with two types of encoding in the proposed model: \romannumeral 1) Encoder model \romannumeral 2) Autoencoder model, as given in Figures~\ref{fig:encoder} and~\ref{fig:autoencoder} respectively. For encoder model we use two linear layers and one ReLU layer. Similarly for the autoencoder model we use two linear layer and one ReLU layer in both encoder and decoder part. As given in Algorithm~\ref{alg:topic_representation}, we extract weighted topic representation $\tau_{n_j}$ using our knowledge weight variable $(1-\beta)$.

\subsubsection{Knowledge representations $\lambda_{n_j}$}
We propose the knowledge representations of $n_j$ as the weighted aggregation of Wikipedia representations ($\omega_{n_j}$) and presidential debates representations ($\tau_{n_j}$). We use the weight variable $\beta \in [0,1]$ to optimize the importance of representations from two different knowledge bases as given in Equation~\ref{eq:Know_rep}.

\begin{equation}
\lambda_{n_j} = [\beta \times \omega_{n_j}] \bigoplus [(1-\beta) \times \tau_{n_j}]
\label{eq:Know_rep}
\end{equation}

\subsection{News Article Representations $\Theta$}
With base representations $\delta$ learned from a news article and its corresponding knowledge representations $\lambda$, we extract the knowledge-infused news article representations as $\Theta = \delta \bigoplus \lambda$. We further fine-tune all three representations of news articles ($\delta$, $\omega$, and $\tau$) by training a supervised model $f(\Theta)$ to predict political leaning of a news article.

\subsection{Training the Proposed Model}
The concatenated knowledge-infused  representations $\Theta$ captures the high level information. Then $\Theta$ is fed to a fully connected linear layer followed by a ReLU layer as shown in Equation \ref{eq:relu} to predict the political leaning of news articles. Here \emph{W} is parameter matrix and \emph{b} is the bias associated.

\begin{equation}
Pr(\mathcal{C}_{n_j}|\Theta)= ReLU(W\Theta+b)
\label{eq:relu}
\end{equation}
We optimize the knowledge-infused news article representations for the supervised task using the multi-class cross entropy loss function, as given in Equation~\ref{eq:loss}, to optimize model parameters. 
\begin{equation}
\mathcal{L}= - \sum_{\mathcal{C}_{n_j}=1}^{M} y_{\mathcal{C}_{n_j,\Theta }}   log(Pr(\mathcal{C}_{n_j}|\Theta))
\label{eq:loss}
\end{equation}

where $\mathcal{C}_{n_j}$ is the training target label with respect to $\theta$, M is the number of classes and y is binary indicator that represents correct or incorrect classification of the class label. In this paper, we set $M = 3$ (\emph{left}, \emph{center}, and \emph{right}).



\begin{table*}[ht]
\centering
\caption{Impact of infusing external knowledge to news article representations in the prediction task. All model results are given in Accuracy(\%), Precision, Recall, Macro F1, and Mean Average Error (MAE). The proposed knowledge-infused approach with encoder module to learn topic representations ($\tau$) gives significant performance boost over the baseline approaches in terms of F1 score.} \label{table:main_result} 
\begin{tabular}{c|p{1.25cm} p{1.5cm} p{1.5cm} p{1.5cm}p{1.5cm}}
 \hline
 \hline
 Experiment Name & Acc. & Precision& Macro F1 & Recall & MAE\\
 \hline
 NewsA(RandomSplit) & 0.749 &  0.7350 & 0.7456 & 0.8145 & 0.3774\\
 \hline
 Baly et al.(2020)(NewsA)~\cite{baly2020detect} & 0.3675 & - & 0.3553 & -& 0.90\\
 Devatine et al.(2022)(Bi-LSTM)~\cite{devatine-etal-2022-predicting} & 0.4697 & - & 0.4441 &-& 0.69\\
 NewsA(Our method) & \textbf{0.516} & \textbf{0.5325} & \textbf{0.4804} & \textbf{0.5319} & \textbf{0.5907}\\
\hline
 WikiA & 0.562 & 0.4722& 0.4040& 0.3650& 0.5625\\
  Baly et al.(2020)(NewsA+WikiA)~\cite{baly2020detect} & 0.4975 & - & 0.5116 & - & 0.32\\
   Devatine et al.(2022)(Bi-LSTM + SA/Sent)~\cite{devatine-etal-2022-predicting} & 0.4876 & - & 0.4584 &-& 0.67\\
   Liu et al.(2018)+SA/EDU~\cite{liu-lapata-2018-learning}& 0.5101 & - & 0.4861 &-& 0.72\\
 Devatine et al.(2022)(Bi-LSTM + SA/EDU)~\cite{devatine-etal-2022-predicting}& 0.5439 & - & 0.5136 &-& 0.57\\

 NewsA+WikiA   &  \textbf{0.685}  & \textbf{0.5535} &   \textbf{0.6290} & \textbf{0.6835} &\textbf{0.6011}\\
\hline
  Khan et al.(2023))~\cite{ko2023khan}& 0.3240 & 0.33 & 0.161 &0.10 & 1.03\\
  Baly et al.(2020)(NewsA+Twitter)~\cite{baly2020detect} & 0.72 & - & 0.6429 & - & 0.2900 \\
 NewsA+WikiA+Topic(AE)  &0.5867 & 0.6067& 0.5344 & 0.5936 & 0.6184\\
  NewsA+WikiA+Topic(E) & \textbf{0.73} & \textbf{0.7232} & \textbf{0.7288} & \textbf{0.7584} & \textbf{0.3676}\\
 \hline
\end{tabular}

\end{table*}

\section{Experiments and Results}

In this section, we evaluate our proposed knowledge-infused model with multiple experimental setup for political leaning detection task. We give results based on multiple transformer-based text representation models, assessed the model fairness with multiple datasets, and analyze the impact of weight parameter $\beta$ on knowledge representations infused with base news article representations $\delta_{n_j}$.

\subsection{Experimental Setup}
Before presenting our experimental results, we will detail the implementation details and notations of the baseline models in the below sections.

\subsubsection{Implementation details}

In this work, we use PyTorch 1.12.1~\cite{10.5555/3454287.3455008} to implement our proposed model. The machine we use for all the experiments is equipped with an Intel X710 quad port CPU with 64.14 GB memory and an NVIDIA Ampere A10 GPU with 24 GB memory, the GPU is installed with CUDA 11.3 and cuDNN8.3.2. We set the batch size as 2 and we use Adam optimizer ~\cite{Kingma2014AdamAM} with learning rate = 1e-6. Lastly, we set the number of epochs as 3.

\subsubsection{Baseline models}

We compare the performance of our proposed model with the performance reported in the following state-of-the-art baselines on the same media split dataset. Since there is not much work in the literature to predict the political leaning of news articles, we use multiple modules in our proposed model as baselines as well. Unless or otherwise specified as \emph{Random Split}, all the models used in our experiments are trained with \emph{Media Based} data split. All the baseline models used in this paper are given below:

\begin{itemize}
\item \textbf{Devatine et al.(2022) (Bi-LSTM)}~\cite{devatine-etal-2022-predicting} This model is their base model and it leverages sequences of static word embedding to generate hidden representations. This model is modified to incorporate the enhancements proposed in ~\cite{ferracane-etal-2019-evaluating} and it achieves $46.97\%$ accuracy.
\item \textbf{Devatine et al.(2022) (Bi-LSTM+SA/Sent)}~\cite{devatine-etal-2022-predicting} This model is an extension of the Bi-LSTM model where it leverages structural attention and sentence segmentation to the existing Bi-LSTM model to give $48.76\%$ accuracy. 
\item \textbf{Devatine et al.(2022) (Bi-LSTM+SA/EDU)}~\cite{devatine-etal-2022-predicting} In particular, the model performs better when they incorporate Elementary Discourse Units(EDU) segmentation(Bi-LSTM + SA/EDU) to the base Bi-LSTM model. Identifying text spans(EDU) is linked with discourse relations to improve the overall model performance.
\item \textbf{Liu et al.(2018) + SA/EDU} The original model proposed in ~\cite{liu-lapata-2018-learning} leverages structural attention and sentence segmentation performs well without the improvements introduced in ~\cite{devatine-etal-2022-predicting}.
    \item \textbf{Baly et al.(2020)(NewsA)~\cite{baly2020detect},
    Baly et al.(2020)
    (NewsA+ WikiA)~\cite{baly2020detect}, and Baly et al.(2020)(NewsA+Twitter)~\cite{baly2020detect}} This work is the closest resemblance to the proposed model. But this work utilizes Twitter follower Bios, which can be uncertain on political leaning, and Wikipedia articles to predict the political leaning of news articles. Also, this work has not given any experiments to evaluate the model fairness on multiple datasets.
  \item \textbf{WikiA}: We report the performance of the proposed model only with Wikipedia articles representation from \textbf{WKB} and not including any base representations $\delta_{n_j}$.
  \item \textbf{NewsA+WikiA}: This is the proposed model which learns the Wikipedia representation $\omega_{n_j}$ together with news article representations $\delta_{n_j}$ to predict the political leaning of the news article $n_j$. 
  \item \textbf{NewsA+WikiA+Topic(E)} and \textbf{NewsA +WikiA+Topic(AE)}: These are the proposed models which utilizes Wikipedia $\omega_{n_j}$ and topic representations $\tau_{n_j}$ with the base news article representations where \emph{E} and \emph{AE} denotes Encoder and Autoencoder versions of the model respectively.
  
\end{itemize}

\subsubsection{Language Models}
We utilize the existing transformer-based text representation learning models in our proposed model. We primarily use BERT~\cite{DBLP:conf/naacl/DevlinCLT19} in all our experiments. However, we also give results by replacing BERT with other popular language models like \emph{RoBERTa} ~\cite{liu2019roberta} and \emph{DistillBERT}~\cite{Sanh2019DistilBERTAD} to learn base representations $\delta_{n_j}$ and Wikipedia representations $\omega_{n_j}$ of a news article $n_j$. 




\subsection{Results}


Table \ref{table:main_result} gives an overview of the results of all our experiments using BERT to obtain base representations $\delta_{n_j}$ and Wikipedia representations $\omega_{n_j}$ of a news article $n_j$. We replicate some of the results in Table~\ref{tab:political_bias} in our results for extensive comparison of our model performance. this experiment, we set $\beta=0.5$ to give more importance to topics and all the results in Table~\ref{table:main_result} are based on the Media-based split(1) dataset. 


We can note from Table~\ref{table:main_result} that the performance of the base BERT model on only news articles differs significantly for the random and media-based data splits. This corroborates our claim that models are not able to generalize the political learning from unseen news domains. It is worth mentioning here that minor parameter tuning on BERT model increase the model accuracy by $15\%$ compared to the existing work~\cite{baly2020detect}. To reduce the model bias we evaluated several experiments by infusing external knowledge sources. 

We first compare the model performance by only utilizing Wikipedia representations $\omega_{n_j}$ to predict the political leaning of a news article $n_j$. Similar to the previous experiment, we notice that the parameter fine-tuning on the BERT model increases the model accuracy by $16.9\%$ compared to the model that uses only base representations while the existing method~\cite{baly2020detect} under performs in a same setup. Also, our proposed model outperforms all other baseline works in terms of performance measures. Our experiments improves the accuracy by $17\%$ than {Liu et al.(2018)+SA/EDU~\cite{liu-lapata-2018-learning}. Furthermore, it increases accuracy by $19.8\%$ and $14\%$ than Devatine et al.(2022) (Bi-LSTM+SA/Sent) and Devatine et al.(2022) (Bi-LSTM+SA/EDU)~\cite{devatine-etal-2022-predicting} respectively, when we utilize Wikipedia representations $\omega_{n_j}$. Next, we compare the proposed model that infuses external knowledge representations $\lambda_{n_j}$ of a news article $n_j$ with the base representations $\delta_{n_j}$, where $\lambda_{n_j}$ is the weighted aggregation of Wikipedia representations $\omega_{n_j}$ from \textbf{WKB} and topic representations $\tau_{n_j}$ from \textbf{PDB}. It is evident from Table~\ref{table:main_result} that the proposed model with \emph{Encoder} setup outperforms the existing work~\cite{baly2020detect} and the \emph{Autoencoder}. The encoder version of the proposed model gives an accuracy of $73\%$ to predict the political leaning of news articles whose news domains are completely invisible during the training phase. Even though the performance of the Auto Encoder version of the proposed model is almost equal to that of the existing work~\cite{baly2020detect}, we emphasize that the proposed model gives a significantly better performance in terms of \emph{F1 score}.

\begin{table}[h!]
\centering
\caption{Impact of varying the importance of representations $\omega_{n_j}$ and $\tau_{n_j}$ with the weight parameter $\beta$ on the proposed model performance} \label{table:abl_2} 

\begin{tabular}{ l| l | l l l l l }
 \hline
 \hline
\multicolumn{2}{c|}{\multirow{2}{*}{Model}} & \multicolumn{5}{c}{$\beta$} \\
\cline{3-7}
\multicolumn{1}{l}{} & & 0.0 & 0.1 & 0.5& 0.7& 1.0\\
 \hline
E & Acc. & 0.5044 & 0.5704& \textbf{0.73} &0.6850 &0.6474\\ 
& Mac. F1&0.4801 & 0.5885 & \textbf{0.7232}& 0.6321 & 0.6452\\
\hline
AE & Acc. &0.5066 &0.5824 & \textbf{0.5867} & 0.4191 & 0.4027\\ 
& Mac. F1&0.4675& 0.5196& \textbf{0.5344} & 0.3906 & 0.3587\\
\hline
\end{tabular}
\end{table}

\begin{table*}[h]
\centering
\caption{The proposed model performance with different backbone language models. Both BERT and RoBERTa models are able to mitigate algorithmic political bias with the proposed knowledge-infusion approach} \label{table:abl_3} 

\begin{tabular}{p{1.75cm}p{3.5cm}| p{1.5cm} p{1.5cm} p{1.5cm} p{1 cm}p{1cm}}
 \hline
 \hline
\multicolumn{2}{c|}{\multirow{1}{*}{Model}} & Acc. (\%) & Precision& Macro F1 & Recall & MAE \\
\hline
RoBERTa & (NewsA) & 0.361 & 0.333& 0.176 & 0.120 & 0.963\\
&NewsA+WikiA+Topic(E)& 0.584& 0.607& 0.510& 0.710&0.637\\
&NewsA+WikiA+Topic(AE) & 0.525 & 0.544 & 0.451& 0.672 & 0.511\\
\hline
DistilBERT & (NewsA) & 0.497 & 0.511 & 0.470 & 0.497 & 0.670\\
&NewsA+WikiA+Topic(E) &0.323& 0.333& 0.163& 0.107& 1.036\\
&NewsA+WikiA+Topic(AE) &0.323& 0.333& 0.163& 0.107& 1.036\\
 \hline
BERT & (NewsA) & \textbf{0.516} & \textbf{0.532} & \textbf{0.480} & \textbf{0.531} & \textbf{0.590}\\

 &NewsA+WikiA+Topic(AE)  &\textbf{0.5867} & \textbf{0.6067}& \textbf{0.5344} & \textbf{0.5936} & \textbf{0.6184}\\
 
  &NewsA+WikiA+Topic(E) & \textbf{0.73} & \textbf{0.7232} & \textbf{0.7288} & \textbf{0.7584} & \textbf{0.3676}\\
 \hline
\end{tabular}
\end{table*}

In Table \ref{table:abl_2} we show the impact of varying the value of weight parameter $\beta$ on both Encoder and AutoEncoder versions of the proposed model. We notice that both models give their best performance with $\beta = 0.5$, which means we set equal priority to both knowledge sources. We verify the efficacy of our external knowledge bases, and the results illustrate the both global knowledge base and topic based knowledge plays an important role in terms of reducing the algorithmic political bias and improve in predicting the political leaning of news articles.

\begin{figure}[ht]
\centering
\includegraphics[scale=0.3]{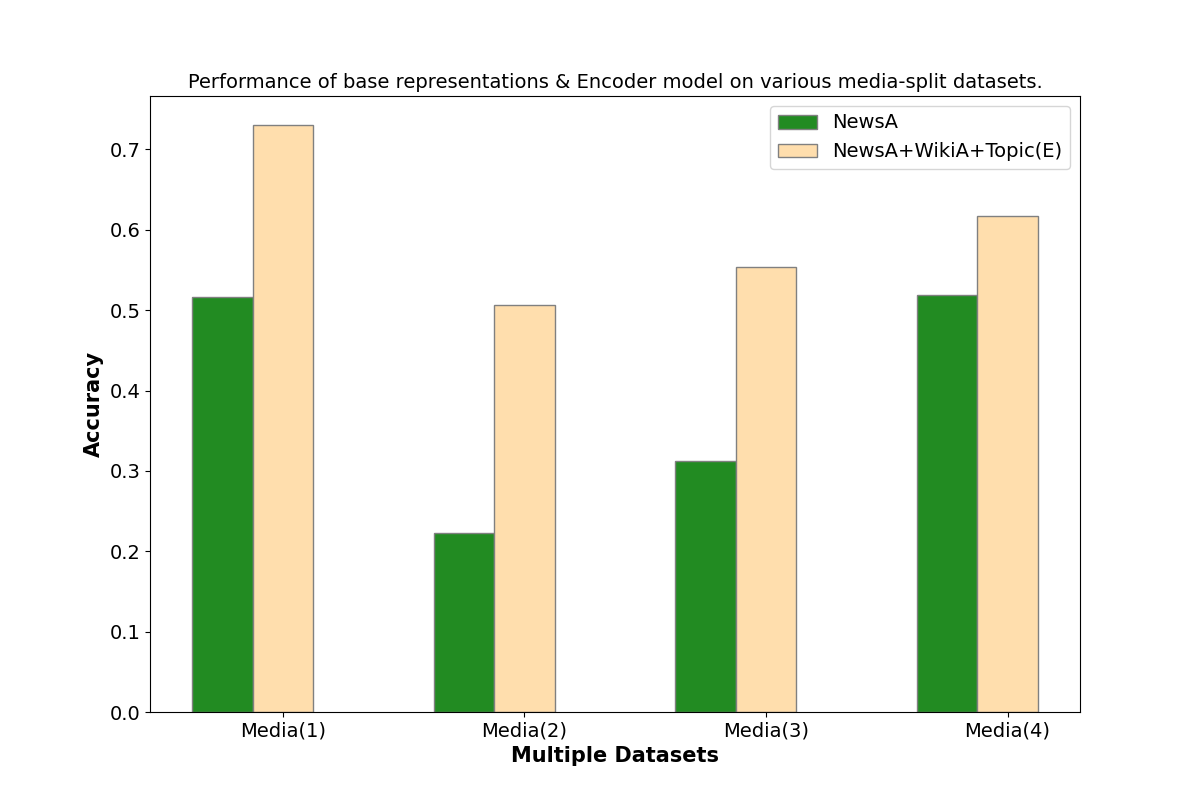}
\caption{Accuracy (\%) of the proposed model with only base representations (\emph{NewsA}) and with knowledge-infused representations (\emph{NewsA+WikiA+Topic(E)}) on multiple media-based data splits. The knowledge-infused approach has a significant performance improvement over the base model in mitigating algorithmic political bias}
\label{fig:diff_datasets}
\end{figure}

In order to demonstrate the generalizing capabilities of the proposed model, we evaluate the proposed model on multiple Media-based data splits. From the Figure~\ref{fig:diff_datasets}, we notice that the algorithmic political bias is present in the base representation without any external knowledge on all the Media-based split datasets. Figure~\ref{fig:diff_datasets} also displays the effectiveness of our proposed framework across multiple train and test set combination from our news article dataset.It is important to analyze the performance of the model by varying the dataset during evaluation. We show the performance of our model on different Media-based splits to illustrate robustness and generalizability of the model. 




To demonstrate the efficacy of our proposed model we use multiple language models like RoBERTa and DistillBERT. Table \ref{table:abl_3} confirms that among all the models, BERT representation model outperforms other language models in both Encoder and AutoEncoder versions. BERT is a strong baseline model, as it was pre-trained on a large corpus of text and has been fine-tuned on various downstream tasks. RoBERTa and DistillBERT are both variations of BERT, RoBERTa, for example, uses additional pre-training data and modified pre-training techniques, while DistillBERT uses a knowledge distillation approach to reduce the computational resources required for training. Although efficient, we noticed that the proposed model trained with DitillBERT gives poor performance and it is unable to mitigate the algorithmic political bias in our prediction task. Table~\ref{table:abl_3} also corroborates the fact that BERT is better suited than RoBERTa or DistillBERT for capturing certain linguistic features or for making political leaning prediction. 








\section{Conclusion and Future Work}
In this work, we attempt to mitigate algorithm political bias of machine learning algorithms by infusing external knowledge sources like Wikipedia and political debates to predict the political leaning of news articles. We proposed a novel way to learn weighted feature representations of entities or topics present in presidential debate records by carefully mapping them in news articles. With a series of experiments, we notice that external knowledge sources can debias base feature representations of news articles and thereby improving the performance of the prediction model by outperforming the baseline approaches in terms of prediction accuracy (\%) and F1 score. We demonstrate the effectiveness of our proposed knowledge-infused model by conducting several experiments on a variety of media-based data splits, and with multiple base language models. 

The proposed model and the problem of predicting the political leaning of news articles have several future directions. We discuss some of them below:


\subsection{Quantifying Political Bias on Topics} The proposed work and existing work characterize political bias on fine-grained entities like users, social media posts, and news articles. An interesting research direction can be identifying and forecasting political bias of topics and events with news articles as surrogate information using machine learning. Such models can quantify social responses to news content on a given topic or story which can help social media engineers to give supporting information on topics for news articles from a given news domain.

\subsection{Mitigating echo chambers} It is well known that the political bias in news domains creates echo chamber effect in social media communities. Identifying the political leaning of news articles can assist analysts to engineer measures to mitigate echo chamber effect in social media and online discussion forums. This will open opportunities to develop ML algorithms that recommend news contents which are similar in story that the user prefers but in different projections of the story. Another potential research direction can be mitigating polarized online discussions among multiple users who argues about political scenarios that appeared on news articles. 

\subsection{Dynamics of polarization in online forums} Since the political leaning of news domains change over time, mitigating the political bias in a temporal aspect can also be another interesting future direction in this work. Understanding the dynamics of political bias of topics in online forums can also be another potential research direction given the importance of quantifying political bias of topics.


\bibliographystyle{IEEEtran}
\bibliography{bibliography}

\end{document}